%% file: main.tex
\documentclass{article}
\usepackage{spconf,amsmath,graphicx}

\input{src/header}

\title{On the Query Strategies for Efficient Online Active Distillation}
\name{Michele Boldo$^{\star}$ \hspace{0.2cm} Enrico Martini$^{\star}$ \hspace{0.2cm} Mirco De Marchi$^{\star}$ \hspace{0.2cm} Stefano Aldegheri$^{\dagger}$ \hspace{0.2cm} Nicola Bombieri$^{\dagger}$}
\address{$^{\star}$ Department of Computer Science, University of Verona, Verona, Italy. \\
      $^{\dagger}$ Department of Engineering for Innovation Medicine, University of Verona, Verona, Italy.}
\begin{document}
%
\maketitle
\begin{abstract}
\input{src/abstract}
\end{abstract}
\begin{keywords}
Online Distillation, Active Learning, Human Pose Estimation, Run-time Model Adaptation, Edge Training
\end{keywords}

\section{Introduction}
\input{src/introduction}

\section{Background and Related Works}
\input{src/related_work}

\section{Query Strategies}
\input{src/methodology2}

\section{Experimental Results}
\input{src/experiments}

\section{Discussion}
\input{src/discussion}

\section{Conlusion and Future Work}
\input{src/conclusion}

\bibliographystyle{IEEEbib}
\bibliography{bibliography}

\end{document}

%% file: src/header.tex
\usepackage{times}
\usepackage{epsfig}
\usepackage{graphicx}
\usepackage{amsmath}
\usepackage{amssymb}
\usepackage{multirow}
\usepackage{booktabs}
\usepackage{mathtools}
\usepackage{subcaption}

\ifdefined\draft
    \usepackage{comment}
    \usepackage[textsize=footnotesize]{todonotes}
    \DeclareRobustCommand{\enrico}[1]
    {{\todo[color=green!40,inline]{EM: #1}}} 

    \DeclareRobustCommand{\michele}[1]
    {{\todo[color=orange!40,inline]{MB: #1}}} 
    
    \DeclareRobustCommand{\nicola}[1]
    {{\todo[color=yellow!40,inline]{NB: #1}}} 

    \DeclareRobustCommand{\mirco}[1]
    {{\todo[color=brown!40,inline]{MDM: #1}}}

    \usepackage[nopar=false]{lipsum}
    
    \DeclareRobustCommand{\fake}[1]{\color{gray!40!}{\lipsum[][#1]}\color{black}}
    
\else 
\usepackage{pbalance} 
\newcommand{\enrico}[1]{}
\newcommand{\nicola}[1]{}
\newcommand{\michele}[1]{}
\newcommand{\mirco}[1]{}
\newcommand{\fake}[1]{}

\fi

\usepackage[pagebackref=true,breaklinks=true,letterpaper=true,colorlinks,bookmarks=false]{hyperref}


%% file: src/abstract.tex

Deep Learning (DL) requires lots of time and data, resulting in high computational demands. Recently, researchers employ Active Learning (AL) and online distillation to enhance training efficiency and real-time model adaptation. This paper evaluates a set of query strategies to achieve the best training results. It focuses on  Human Pose Estimation (HPE) applications, assessing the impact of selected frames during training using two approaches: a classical offline method and a online evaluation through a continual learning approach employing knowledge distillation, on a popular state-of-the-art HPE dataset. The paper demonstrates the possibility of enabling training at the edge lightweight models, adapting them effectively to new contexts in real-time.

%% file: src/introduction.tex
Deep Learning (DL) techniques use large amounts of annotated data and high computing times \cite{Schwartz2020}. To address this, researchers have employed techniques to improve training efficiency \cite{Mehlin2023}, including the use of \textit{active learning (AL)}. The goal of AL is to train a neural network on the most significant samples to achieve high accuracy while minimizing computation time.
\textit{Online active learning} is a rising derivation of AL, suitable for handling dynamic data streams that arrive sequentially over time. It selects a subset of samples from a video source to use in training, making it well-suited for real-time and resource-constrained environments. However, obtaining accurate labels quickly in real-time scenarios can be challenging and impact the approach's effectiveness. \textit{Online distillation}, a continuous learning technique where a ``teacher" model transfers knowledge to a ``student" model in real-time as new data arrives, can address this challenge. Integrating these methods offers several advantages, such as maximizing knowledge transfer and reducing the need for extensive labeling efforts while ensuring real-time model adaptation to new data.

This paper presents an experimental study addressing two fundamental questions. Firstly, determining the amount of data required for the neural network to achieve sufficient generalization and high accuracy when applied to human pose estimation (HPE).  Secondly, identifying the criteria to select the subset of frames. 
\begin{figure}[t]
\centering
\begin{subfigure}[t!]{.95\linewidth}
\includegraphics[width=\linewidth,page=1,trim={0 1cm 0 0},clip]{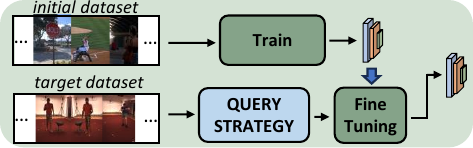}
\caption{Offline fine tuning.}
\label{fig:offline_training}
\end{subfigure}
\begin{subfigure}[t!]{.95\linewidth}
\includegraphics[width=\linewidth,page=2]{fig/fig1_v2.pdf}
\caption{Online active distillation framework.}
\label{fig:online_training}
\end{subfigure}
\caption{}
\label{fig:intro_figures}
\end{figure}
The analysis focuses on HPE, which estimates the locations of human body joints from images or videos. HPE finds applications in various domains, from clinical \cite{martini2022enabling} to industrial \cite{lim2021} settings, with variations introduced by factors such as camera orientation and lighting conditions. Fine-tuning is often necessary to ensure optimal performance across different scenarios. 

In the context of AL, a \textit{query strategy} refers to the approach used to select which frames from a dataset should be added to the training set. To evaluate the impact of selected frames during training, our study employs two approaches: a classical offline method and a continual learning-based approach. In the first approach the entire dataset is available to the query strategy, while in the second approach the images are available incrementally, and the dataset on which the strategy operates is continuously changing over time, demonstrating the effectiveness of the strategies in a real scenario. Figure \ref{fig:offline_training} shows the offline fine tuning framework, where 
a lightweight model, initially pre-trained on a large generic dataset, is further fine-tuned using only a subset of a specific target dataset. Figure \ref{fig:online_training} illustrates an Edge AI online learning framework. In this setup, a device with limited computational resources conducts real-time inference and training on a video stream simultaneously. The soft labels are provided by a larger model that achieves higher accuracy at a lower frame-rate. To ensure efficient training on the embedded device, the query strategy selectively picks only essential data. In both frameworks, we test four different query strategies: uniform, random, error-based, and uncertainty-based.

The contributions are the followings:
\begin{itemize}
    \item Evaluation of query strategies to achieve data efficiency in classical offline fine-tuning of a lightweight model.
    \item Evaluation of query strategies for continual learning context based on knowledge distillation for online domain adaptation on low-powered devices.
\end{itemize}
The evaluation is performed on a popular state-of-the-art HPE dataset, enabling edge training with the adaptation of a generic lightweight model to a new context in real-time.

\michele{Dire: è risputo dall'esperienza che la qualità di un train degrada al diminuire dei sample, ma to the best of our knowledge nessuno ha mai quantificato ciò }

%% file: src/related_work.tex
In situations where there is a constraint on the number of samples that can be used for training, AL is used to select a subset of the initial dataset~\cite{Xu2021}. Lu et al.~\cite{lu2023rebenchmarking} proposed a benchmark the performance of AL strategies for binary classification. 
In the context of HPE, Liu et al.~\cite{liu2017active} proposed an AL framework to reduce the amount of annotations required among a large unlabelled dataset. They employed a sampling strategy based on the uncertainty that trains the network only in frames where the HPE model is not confident. Yoo et al.~\cite{yoo2019learning} developed a loss prediction model that learns to limitate the loss defined in
the target model. Zhang et al.~\cite{Zhang2021} proposed a query strategy that 
selects frames balancing between information representativeness and uncertainty.

In the HPE context, Zhang et al.~\cite{Zhang_2019_CVPR} trained a small CNN using labeled data and a larger teacher model. Hwang et al.~\cite{Hwang2020} proposed \textit{MoVNect}, a lightweight HPE framework for real-time mobile processing, using soft labels from a teacher. \textit{Online distillation} is a rising variant of KD that involves continuously updating a student model with new knowledge from the teacher in real-time as new input data are encountered. Cioppa et al.~\cite{Cioppa2019} used online distillation for real-time human segmentation, showing the adaptive capability of the network.
Mullapudi et al.~\cite{Mullapudi2019} proposed \textit{Just-In-Time}, a framework that performs video semantic segmentation on low-cost devices. Khani et al.~\cite{Khani2021} proposed \textit{Adaptive Model Streaming (AMS)}, a framework for video semantic segmentation to reduce network bandwidth by adaptive sampling. Houyon et al.~\cite{Houyon2023} addressed two challenges of online distillation (i.e. domain shifts and catastrophic forgetting) in semantic video segmentation using a replay-based method.

Online AL is an extension of AL for data that are available incrementally. By selecting a subset of the data, it is efficient in terms of resource utilization, while maintaining accuracy performance over time even in presence of changes of the data distribution, making it useful in scenarios where quick and timely responses are crucial.
Cacciarelli et al.~\cite{Cacciarelli2023} proposed an overview of the most recent approaches for selecting informative observations from streams of data.
Manjah et al.~\cite{manjah2023stream} proposed \textit{StreamBased Active Distillation (SBAD)}, a real-time online AL framework to fine-tune a light-weight model that performs object detection in videos.

To the best of our knowledge, this is the first paper that investigate on which are the most effective query strategies for online AL applied to HPE.

%% file: src/methodology2.tex

The primary goal of AL is to efficiently select the most informative data frames for annotation, thereby reducing the labeling effort while maintaining or improving the model's performance. In this context, we explore four query strategies used to fine-tune a lightweight HPE model.
Let's denote the set of data samples as $\mathcal{D}$. A query strategy is a function that selects a subset $\mathcal{D}' \subseteq \mathcal{D}$ of images, based on a specific criterion or rule. 
All query strategies employed in this paper are implemented as fixed-rate strategies, where they operate by selecting a specific percentage of frames within a defined window of samples. 
We report the definitions for each of the query strategy tested:

\begin{itemize}
 \item \textit{Uniform sampling}: 
This is a simple strategy that evenly selects frames within a defined window at a fixed rate. 
Given a set of images $\mathcal{D} =  [I_0 \dots I_N ]$ and a sample rate $p \in (0,1)$, the subset selected is $\mathcal{D}'=\{I_i:i= \lfloor j / p \rfloor, j\in [0, \lfloor N\cdot p\rfloor)\}$.
It ensures a balanced representation of data but may be less efficient in identifying informative samples compared to other strategies. It works well when informative samples are uniformly distributed, but may not prioritize critical instances efficiently, especially in imbalanced datasets. 

\item \textit{Random sampling}: 
It selects frames randomly within the designated window. Given a set of images $\mathcal{D} =  [I_0 \dots I_N ]$ and a sample rate $p \in (0,1)$, the subset $\mathcal{D}'$ is selected by random sampling $|\mathcal{D}'|=\lfloor N\cdot p \rfloor$ frames. It is simple to implement and unbiased. However, it may include less informative frames and miss critical instances, slowing down learning.

\item \textit{Error-based sampling}: 
In this strategy, frames that result in larger errors during the model's inference are chosen. In~\cite{yoo2019learning}, this approach is implemented as the ``Loss Prediction Module", a model that estimates the model's loss on an image. To assess the effectiveness of this technique, we introduced an ``oracle" model that precisely selects all images where the error is higher. Given a set of images $\mathcal{D} =  [I_0 \dots I_N ]$ and a sample rate $p \in (0,1)$, the subset $\mathcal{D}'$ is composed by the $|\mathcal{D}'|=\lfloor N\cdot p \rfloor$ frames that produce the largest error compared to the ground truth.

\item \textit{Confidence-based sampling}:
This strategy selects frames where the model is less confident in its predictions. Given a set of images $\mathcal{D} =  [I_0 \dots I_N ]$ and a sample rate $p \in (0,1)$, the subset $\mathcal{D}'$ is composed by the $|\mathcal{D}'|=\lfloor N\cdot p \rfloor$ frames that have the lowest peak in the heatmap across all keypoints.

\end{itemize}

%% file: src/experiments.tex

\begin{figure*}[t]
\centering
\begin{subfigure}[t!]{\linewidth}
\centering
\includegraphics[width=\linewidth]{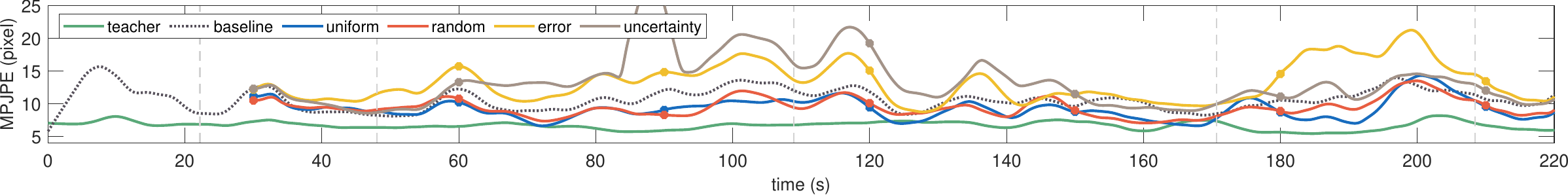}
\caption{Comparison between query strategies, all with a sampling rate of 1\%.}
\label{fig:sampling_methods}
\end{subfigure}
\begin{subfigure}[t!]{\linewidth}
\centering
\includegraphics[width=\linewidth]{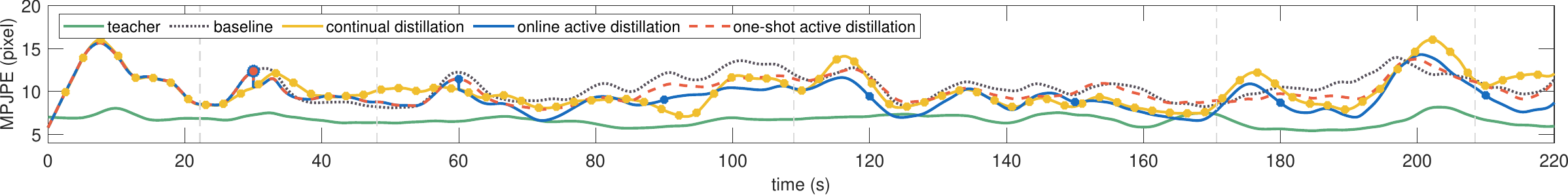}
\caption{Comparison between training rates.}
\label{fig:train_frequency}
\end{subfigure}
\caption{Effect of different sampling techniques on the accuracy of the online active distillation framework. Each grey vertical line corresponds to the beginning of an unseen sequence and each dot corresponds to a weights update.}
\label{fig:plots}
\end{figure*}



\subsection{Implementation details.}
We use a \textit{Densenet} \cite{huang2017densely} CNN model for all the experiments, leveraging the \textit{Nvidia TRTpose Framework}~\cite{trtpose2} for both training and inference.  
We conduct the pre-training and the complete offline fine-tuning process on a standard PC equipped with an Intel i5 7400F processor, 16 GB of RAM, and an Nvidia 2070 SUPER GPU. To test the online active distillation framework, we carry out validation on an Nvidia Jetson Xavier NX. The teacher in this case is Openpose~\cite{openpose}, a HPE framework that relies on a pre-trained CNN. This model provides high accuracy but demands significant computational resources, making it unsuitable for edge devices. 
We use a query strategy to select whether to keep the frames for the next training phase. The training consists of 10 epochs, with a learning rate of $10^{-3}$ and \textit{Adam} optimizer. After completing the training, we optimize the model's weights using the \textit{Nvidia TensorRT Framework}~\cite{torch2trt}. For evaluating the accuracy of the different query strategies, we adopt the \textit{mean per joint position error (MPJPE)}, which is expressed in pixels.

\subsection{Datasets}
For both the experiments, we pre-train the student model using \textit{COCO-Pose}~\cite{lin2015microsoft}, a dataset for training and evaluating deep learning models in keypoint detection and pose estimation tasks. We evaluate the models accuracy on \textit{Human3.6M}~\cite{Ionescu2014}, a widely adopted benchmark for HPE. 
The ground truth is a marker-based MoCap system. For the quantitative analysis of the offline fine-tuning, we utilize actions performed by subjects 1, 5, 6, 7, and 8 to fine-tune the models. Subsequently, we evaluate all methods using the first video of each action performed by subjects 9 and 11. For the experiment of the online active distillation framework, we concatenate all actions performed by subject 1, resulting in a video lasting approximately 21 minutes. The training window is set to 1500 frames (30 seconds), so every updated model is tested on the subsequent training window, before being replaced by the newest one, consistently with the experiments conducted in~\cite{Mullapudi2019,Khani2021}.

\subsection{Results}

In this paragraph we analyze the query strategy techniques in the case of offline fine-tuning. In addition by varying the types of frame considered, we also varied the number of samples, namely taking the 1\%, 5\%, 10\%, 20\% and 40\% of the training dataset. Table \ref{tab:offline-kd} reports the accuracy results, also comparing the frame selection techniques with the model without fine-tuning (i.e. 0\%) and the model fine-tuned on 100\% of the training set. 
\begin{table}[h]
\centering
\caption{Accuracy in MPJPE (pixels) obtained with different sampling rates and query strategies.}
\label{tab:offline-kd}
\resizebox{\linewidth}{!}{%
\begin{tabular}{lc|ccccc|c}
\toprule
& \textbf{0\%}& \textbf{1\%}  & \textbf{5\%} & \textbf{10\%} & \textbf{20\%} & \textbf{40\%} & \textbf{100\%} \\
\midrule
uniform              & 16.33  &  11.92 & \textbf{10.49}  & 10.56  & \textbf{9.92}  & 11.52  & 10.14  \\ 
random               & 16.33  & 11.87  & 10.73 &  \textbf{10.48} &  10.35 & \textbf{9.81}  & 10.14  \\
error            & 16.33  &  12.48 & 11.91  &  10.99 &  10.58 & 10.53  & 10.14  \\
uncertainty       & 16.33  & \textbf{11.82}  &  10.87  & 10.57  & 11.13  &  10.09 & 10.14  \\
\bottomrule
\end{tabular}%
}
\end{table}
In this paragraph, we conduct an extensive analysis on diverse sampling methods employed within the context of online active distillation. Table~\ref{tab:online-kd} reports the accuracy results varying the percentage of the dataset utilized and the teacher model. 
From the columns analyzing the variation in the size of the training dataset with the Openpose teacher, there is not a major improvement training on 1\% or 20\% of the dataset, and the most promising metrics are the random and max-error selection approach. We also substitute the soft label of the teacher with the ground truth to exclude Openpose error (MPJPE of 7.87 pixels on average). As expected, training using ground truth leads to more accurate predictions.
\begin{table}[ht]
\centering
\caption{Accuracy in MPJPE (pixels) obtained varying sampling rate and methods on the online active framework.}
\label{tab:online-kd}
\resizebox{0.96\linewidth}{!}{%
\begin{tabular}{lccccc}
\toprule
   \multirow{2}{*}{\textbf{Methods}} & \multirow{2}{*}{\textbf{\# train}} & \multicolumn{2}{c}{\textbf{Ground Truth}} & \multicolumn{2}{c}{\textbf{Openpose~\cite{openpose}}} \\
                    &   & \multicolumn{1}{c}{\textbf{1\%}} & \textbf{20\%} & \multicolumn{1}{c}{\textbf{1\%}} & \textbf{20\%} \\
\midrule
baseline            & 0   & 13.52             & 13.52         & 13.52             & 13.52 \\
continual distillation        & 382 &    11.21               &    11.21          &     11.84              &  11.84  \\
uniform            & 33  & \textbf{10.79}    & 9.09          & 12.22             & 11.52 \\
random              & 33  & 10.98             & 8.91 & \textbf{11.42}    & 11.54 \\
error        & 33  & 13.67             &      \textbf{8.90}         & 15.61             & \textbf{11.14} \\
uncertainty   & 33  & 14.22             &      10.57         & 15.28             & 11.97 \\ 
\bottomrule
\end{tabular}%
}

\end{table}
Figure \ref{fig:sampling_methods} depicts the progression of errors during training across different sampling metrics over time. The green line represents Openpose, the black dotted line represents the light model without fine-tuning, and the other lines correspond to the sampling metrics. Both the uniform and random sampling metrics show the best outcomes, whereas the other metrics yield slightly worse results than the baseline. Figure~\ref{fig:train_frequency} presents a comparison of different online distillation methods varying the training rate.
The continual distillation (in yellow) trains the model every 128 frames without sampling.
The red dashed line represents the error when the model is fine-tuned only once on 1\% of the sampling window, while the blue line represents the error when efficiently applying the uniform sampling metric. 
The efficient metric achieves better results compared to the alternative approach, saving computation time since the trainings are significantly fewer.

\subsection{Efficiency Analysis}
In the offline fine-tuning evaluation, as we decreased the size of the training dataset, we successfully reduced the computation time from 15 hours to 6 hours, 3 hours, 80 minutes, and 9 minutes namely for the 40\%, 20\%, 10\%, 5\%, and 1\% of the train dataset.
The online test is performed on all sequences of subject 1 concatenated in lexicographical order, forming a 21 minutes video stream. The embedded device (i.e., Nvidia Jetson Xavier NX) receives the frames at 50 Hz. The continual KD framework trains every 128 frames without sampling, so it spends approximately 7 hours to train the models and thus cannot run in real-time. Using an online active distillation framework, we are able to train every 30 seconds, decreasing the number of trainings from 382 to 33 (Table~\ref{tab:online-kd}).  In particular, the strategy that uniformly samples 1\% of frames spends in total only 4 minutes to train, minimizing also the error.


%% file: src/discussion.tex
Active learning can lead to improved results compared to training the model on the entire dataset. The most effective active learning strategies for both offline and continual training are random and uniform, which consistently yield similar outcomes. Generally, these strategies ensure a balanced subset dataset on average, while other approaches may select frames with incorrect labels and false predictions. For dataset percentages greater than 1\% (i.e., 20\%) in continual training, the error-based strategy shows slightly better performance. However, this strategy requires accurate loss estimation without ground truth access, making it impractical in real-world applications. This conclusion is supported in \cite{yoo2019learning}, where they developed a DL model to estimate the loss. Due to its probabilistic nature, it can be compared to an oracle with random elements. On low-powered devices running the online distillation framework, real-time computation is achieved with a very low sampling rate. Some strategies may lead to lower accuracy compared to the baseline, suggesting that sometimes, it is preferred to avoiding fine-tuning.

%% file: src/conclusion.tex
This paper presented an analysis of the impact of data quantity and selection for fine-tuning a HPE neural network. It presents an evaluation of various data query strategies, both for the classical offline approach, which saves computation time, and for a runtime continual learning approach based on knowledge distillation, enabling training at the edge. As future works we plan to extend these strategies to other tasks, such as classification, detection, and segmentation. We also plan to investigate on other strategies as well as the application of other knowledge distillation techniques.

